\documentclass[10pt,twocolumn,letterpaper]{article}

\usepackage{iccv}
\usepackage{times}
\usepackage{epsfig}
\usepackage{graphicx}
\usepackage{amsmath}
\usepackage{amssymb}
\usepackage{booktabs}
\usepackage{subcaption}
\usepackage[final]{pdfpages}
\usepackage[misc]{ifsym}
\usepackage{lipsum}
\usepackage{hyperref}

\iccvfinalcopy 

\def\iccvPaperID{7246}

\ificcvfinal\fi

\newcommand*{\affaddr}[1]{#1} 
\newcommand*{\affmark}[1][*]{\textsuperscript{#1}}
\newcommand*{\email}[1]{\texttt{#1}}
\newcommand\blfootnote[1]{
  \begingroup
  \renewcommand\thefootnote{}\footnote{#1}
  \addtocounter{footnote}{-1}
  \endgroup
}
\begin{document}

\title{Boosting Image Outpainting with Semantic Layout Prediction}

\author{%
$\text{Ye Ma}^*$\affmark[1]\quad $\text{Jin Ma}^{*}$\affmark[2]\quad Min Zhou\affmark[1]\quad Quan Chen\affmark[1]\quad Tiezheng Ge\affmark[1]\quad Yuning Jiang\affmark[1]\quad $\text{Tong Lin}^\dag$\affmark[2]\\
\affaddr{\affmark[1]Alibaba Group} \quad \affaddr{\affmark[2]Peking University}\\
\email{\small\{maye.my,yunqi.zm,chenquan.cq,tiezheng.gtz,mengzhu.jyn\}@alibaba-inc.com}\\
\email{\small\{ma\_jin,lintong\}@pku.edu.cn}
}

\maketitle

\ificcvfinal\thispagestyle{empty}\fi

\blfootnote{${}^*$\; Equal contribution.}
\blfootnote{\dag\; Corresponding author.}

\begin{abstract}
   The objective of image outpainting is to extend image current border and generate new regions based on known ones. Previous methods adopt generative adversarial networks (GANs) to synthesize realistic images. However, the lack of explicit semantic representation leads to blurry and abnormal image pixels when the outpainting areas are complex and with various objects. In this work, we decompose the outpainting task into two stages. Firstly, we train a GAN to extend regions in semantic segmentation domain instead of image domain. Secondly, another GAN model is trained to synthesize real images based on the extended semantic layouts. The first model focuses on low frequent context such as sizes, classes and other semantic cues while the second model focuses on high frequent context like color and texture. By this design, our approach can handle semantic clues more easily and hence works better in complex scenarios. We evaluate our framework on various datasets and make quantitative and qualitative analysis. Experiments demonstrate that our method generates reasonable extended semantic layouts and images, outperforming state-of-the-art models.
\end{abstract}

\section{Introduction}

\begin{figure}[t]
\begin{center}
  \includegraphics[width=\linewidth]{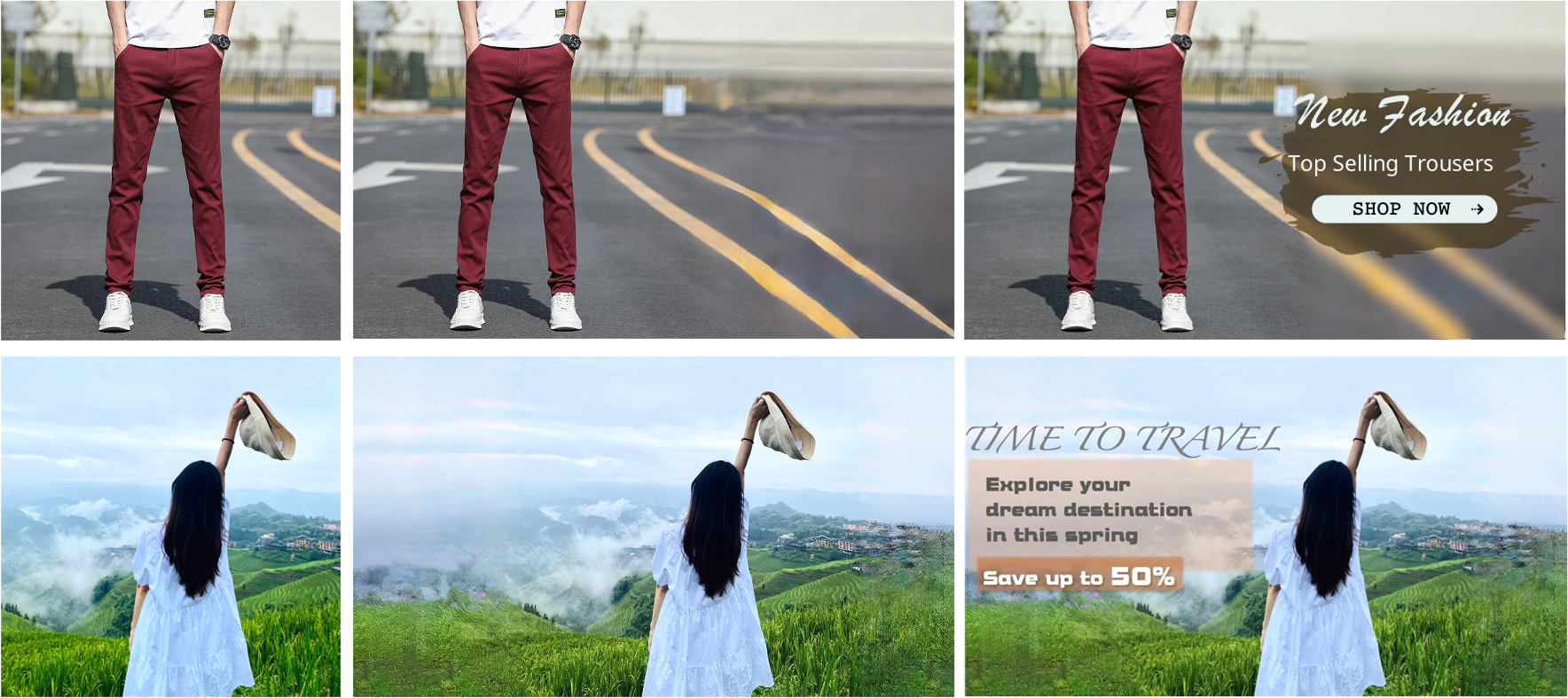}
\end{center}
  \caption{Image outpainting and its applications. It takes input images (left) and generates extended ones (middle). And then designers could create banners (right) based on extended images. Outpainting offers designers the potential to create images of any sizes as they wish.}
\label{fig:intro}
\end{figure}

Image outpainting, also known as image extension, illustrated in Fig.\ref{fig:intro}, aims to extend image boundaries by generating new visual harmonious contents according to the original images.
It could be widely used in applications, such as virtual reality, automatic creative image and video generation, to create more impressive experiences for consumers.

It is obvious that image outpainting is one of the conditional image synthesis tasks~\cite{DBLP:journals/corr/MirzaO14}, whose condition is the original image content.
One similar task which has been usually mentioned with image outpainting is image inpainting~\cite{siggraph00,patchmatch,ContextEncoder,NIPS2015_daca4121,Yu_2018_CVPR,Yu_2019_ICCV,Face_Completion}. 
Though both of them generate unknown regions under the conditions of known regions, outpainting is usually much more difficult.
Because outpainting could only utilize one-side boundary conditions from the original images, whereas inpainting could utilize visual context surrounding the inpainting area in all directions.
Hence, many state-of-the-art image inpainting models have been demonstrated poor performance for image outpainting ~\cite{teterwak2019boundless}.

To solve the image outpainting problem, traditional methods~\cite{image_completion,BiggerPicture,Zhang_2013_CVPR} utilize sophisticated handcrafted features and patch-based matching algorithms to fill in the unknown regions. For the lack of deep semantic features, these methods perform poorly on image extension of complex structure.
To overcome these drawbacks, Wang~\etal~\cite{wang2019wide}, Teterwak~\etal~\cite{teterwak2019boundless} and Yang~\etal~\cite{yang2019very} are the first to design effective one-stage deep models with adversarial training strategies~\cite{goodfellow2014generative} to improve outpainting quality impressively. It demonstrates that high-level semantic representations from deep neural networks play a crucial role in image outpainting.

However, these methods simply model and utilize high-level semantics implicitly learned from CNN architectures from the perspective of general conditional image synthesis. It is far from taking full advantage of the important guiding role of high-level semantics in image outpainting. In this paper, we argue that the explicit prediction of reasonable semantic layout is beneficial to image outpainting in addition to the original image content.

Inspired by the recent prosperity of image segmentation~\cite{deeplabv3,farhadi2018yolov3,zhang2020resnest,tao2020hierarchical} and GAN-based semantic image synthesis~\cite{li2020bachgan,park2019semantic,tang2020local}, we propose a two-stage framework boosting image outpainting with reasonable semantic layout prediction.
Specifically, we decompose the image outpainting task into two generative stages in semantic layout domain and in image domain respectively.
In the first stage, we parse the original known image region into semantic layout with the aid of state-of-the-art segmentation networks. Then we redefine the objective as generating the reasonable layout in unknown regions under the condition of original image content and semantic layout in known regions.
In the second stage, we exploit the extended full semantic layout as auxiliary conditions with image content in multi-scale to synthesize the final realistic image.

Compared with preceding deep generative methods~\cite{wang2019wide,yang2019very,teterwak2019boundless}, our two-stage generative framework has the following advantages:

Firstly, the divide-and-conquer design makes outpainting much easier. The first stage only takes responsibility for the full semantic layout prediction. It is much easier than extending the real image from blank when no texture, color, light or other high frequency information need to be considered. Hence the model could focus mainly on low frequency context instead of high-level representations such as classes, sizes and positions.
Benefiting from the semantic layout of the first stage, the second stage model focuses more on the synthesis of high frequency textures and structures, which makes the learning phase much more stable and the result much more reasonable for complex scenarios.

Secondly, the explicit semantic layout prediction result could provide a handle to improve the quality and diversity of image outpainting results, which is important for interactive image editing applications.

To prove the effectiveness of our proposed method, we conduct experiments on challenging datasets including the ADE20K~\cite{zhou2017scene} and Cityscapes~\cite{Cordts2016Cityscapes}, which contain real world images of complex scenes. Extensive experiments demonstrate that our method could generate images of significantly higher quality in complex scenarios compared to state-of-the-art outpainting models under both quantitative and qualitative assessments. Our contributions are as follows:

(1) We propose a novel two-stage approach to the task of image outpainting. The first stage aims at extending semantic segmentation of known regions to unknown regions, while the second one aims at conditional image synthesis based on the extended segmentation map.

(2) We introduce adversarial losses for both stages and design an encoder-decoder framework to integrate image and semantic segmentation smoothly and subtly.

(3) We apply image outpainting in complex and challenging scenarios, containing objects and backgrounds of various categories. Extensive experiments show our method outperforms state-of-the-art ones by a large margin.

\newcommand{\Paragraph}[1]{\noindent\textbf{#1}~~}

\section{Related Work}

\Paragraph{Generative adversarial networks} In recent years, generative adversarial networks (GANs)~\cite{goodfellow2014generative} are successfully used in many computer vision tasks, such as image synthesis~\cite{biggan,park2019semantic}, image super-resolution~\cite{super_resolution}, image denoising~\cite{denoising} and 3D reconstruction~\cite{slam}. Among these applications, many researchers focus on adversarial loss (WGAN-GP~\cite{wgan_gp}, hinge loss~\cite{hinge_loss}) and normalization structures (SPADE~\cite{park2019semantic}), which can greatly improve training stability and image quality, achieving state-of-the-art for many image generation tasks.

\begin{figure*}[t!]
\centering
\centerline{\includegraphics[width=\textwidth]{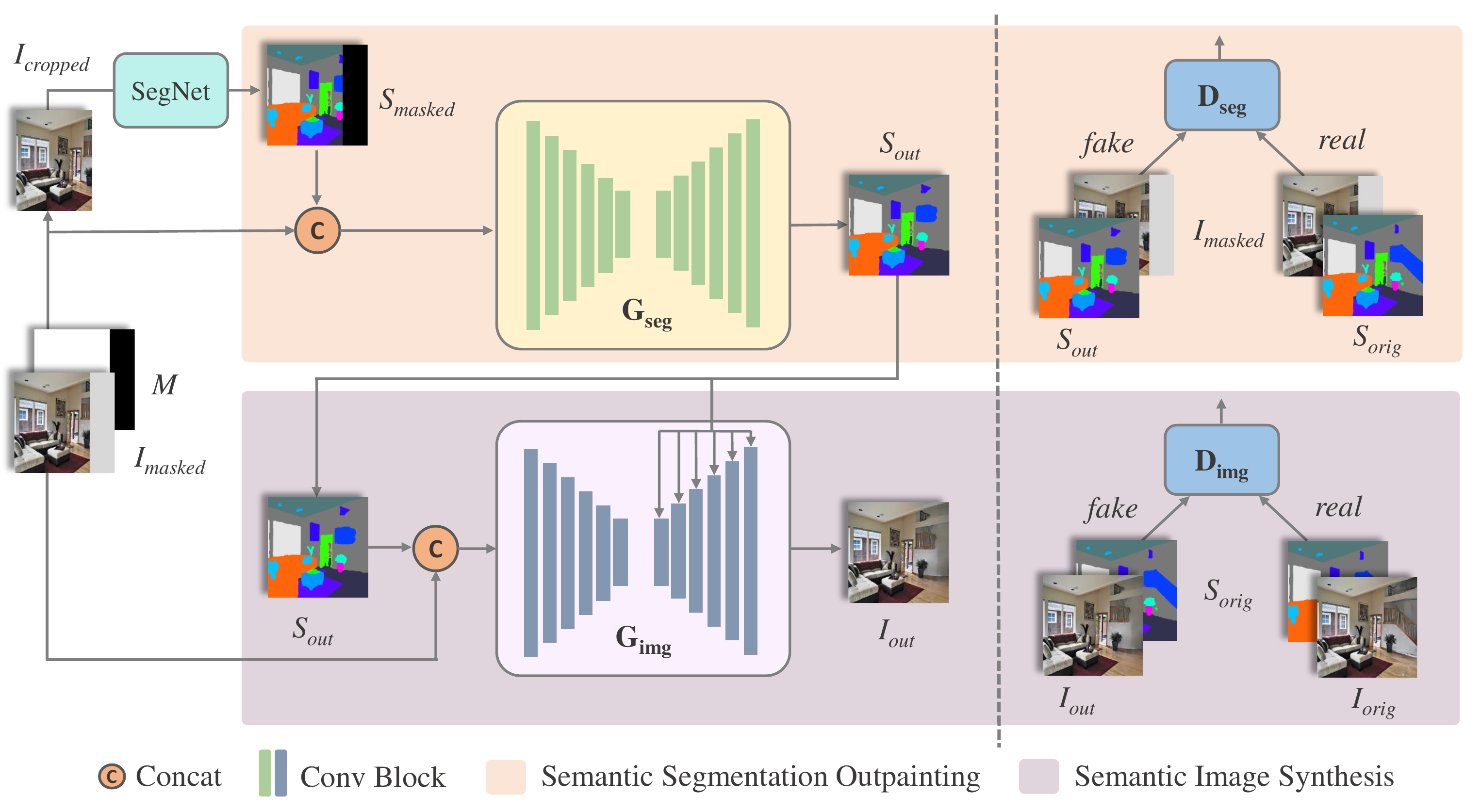}}
\caption{Overview of our two-stage framework. The first generator $\mathbf{G_{seg}}$ exploits masked image $I_{masked}$ and segmentation map $S_{masked}$ to generate extended one $S_{out}$, while the second generator $\mathbf{G_{img}}$ utilizes $S_{out}$ of multi-scale and generates full image $I_{out}$ from masked one $I_{masked}$. The adversarial losses for two stages are on the right.}
\label{fig:overview}
\end{figure*}

\Paragraph{Image inpainting} Previous works~\cite{ballester2001filling,siggraph00,efros1999texture,efros2001image,patchmatch,image_completion} of image inpainting adopt classical computer vision algorithms, which usually search patch candidates in a patch pool for the best similarity and ``paste'' patches to inpainting regions. Various algorithms for constructing patch pool, searching for best candidates and pasting patches consistently are proposed~\cite{siggraph00,patchmatch}. They work decently in simple cases, especially when inpainting areas follow a repeatable pattern. Recently, many researchers adopt end-to-end deep learning models~\cite{iizuka2017globally,ContextEncoder,NIPS2015_daca4121,Yu_2018_CVPR,Yu_2019_ICCV,Face_Completion}. Among them, conditional GANs prove to be effective when handling complex inpainting areas. 
Yu \etal\cite{Yu_2018_CVPR} designs a coarse-to-fine framework to generate high-quality inpainting results with contextual attention, which extracts feature of known surrounding areas to help the generation of unknown areas. Iizuka \etal\cite{iizuka2017globally} employs global and local context discriminators to distinguish real images from synthesis ones to make results both locally and globally consistent.

However, most of the inpainting approaches become unstable and inconsistent when there are large-scale inpainting areas, which occurs usually for outpainting tasks. Inpainting mainly focuses on using information of known region, while outpainting mainly focuses on generating information of unknown area.

\Paragraph{Image outpainting} Similar to image inpainting, early image outpainting models adopt searching and pasting methods~\cite{image_completion,BiggerPicture,Zhang_2013_CVPR}. With the recent development of GANs, a few researchers utilize GANs into image outpainting tasks. Wang \etal~\cite{wang2019wide} propose a feature expansion and context prediction network for image extrapolation. Yang \etal~\cite{yang2019very} adopt LSTM for spatial content prediction and evaluate their encoder-decoder pipeline on a collected scenery dataset. Similarly, Teterwak \etal~\cite{teterwak2019boundless} propose an encoder-decoder model and add pre-trained feature conditions to discriminator for better image quality. These methods use GANs for image outpainting and work decently on scenery photos. However, the lack of semantic segmentation leads to bad performance in complex scenes with objects of various categories.

\Paragraph{Semantic image synthesis} Semantic image synthesis~\cite{li2020bachgan,park2019semantic,tang2020local} aims at converting segmentation masks to real image retaining the semantic information, which is a special form of conditional image synthesis. Park \etal~\cite{park2019semantic} propose a new spatial-adaptive denormalization (SPADE) module, and integrate it to the generator, achieving convincing performance on various complex datasets. 
Hao \etal~\cite{tang2020local} consider learning the scene generation in a local context, and correspondingly design a local class-specific generative network with semantic maps as a guidance, also achieving amazing results.

\section{Methodology}

\subsection{Overview}

Formally, given a ground truth input image ${I_{orig}\in \mathbb{R}^{h \times w \times c}}$ and a binary mask ${M \in \{0,1\}^{h \times w}}$ indicating outpainting region, the model only takes the masked image ${I_{masked} = I_{orig} \odot M}$ as input and generate extended image ${I_{out}\in \mathbb{R}^{h \times w \times c}}$. As shown in Figure~\ref{fig:overview}, we decompose the outpainting task into two stages: semantic segmentation outpainting and semantic image synthesis. During the first stage, the segmentation map $S_{masked}$ and $I_{masked}$ are fed into the semantic segmentation outpainting model to obtain fully extended segmentation map $S_{out}$. In the second stage, semantic image synthesis model focuses on extending texture and color information, taking $S_{out}$ and $I_{masked}$ as input to generate the full extended image $I_{out}$. Following the conventional GAN framework, these two stages both consist of generator and discriminator, and two stages are trained separately.

As for the segmentation map, note that in \textit{NO} phase do we exploit the ground truth segmentation. We adopt a pretrained semantic segmentation model to generate segmentation. Specifically, given a pretrained model $\mathbf{SegNet}$:
\begin{align}
    S_{masked} &= \mathbf{Padding}(\mathbf{SegNet}(I_{cropped})) \\
    S_{orig} &= \mathbf{SegNet}(I_{orig}),
\end{align}
where $I_{cropped}$ refers to the cropped image before outpainting, $\mathbf{Padding}$ refers to right padding zeros and $S_{orig}$ refers to the segmentation map of the whole image.

\subsection{Semantic Segmentation Outpainting}

This part of the task can be treated as a standard outpainting task in semantic domain. The semantic segmentation map contains the complete semantic and layout information of the image, while irrelevant texture and color information is omitted. 

We take the mask $M$, masked image $I_{masked}$ and its corresponding segmentation map $S_{masked}$ as inputs to our generator. The generator $\mathbf{G_{seg}}$ follows common image-to-image translation encoder-decoder structure. The encoder is a fully convolutional network with $11$ conv layers ($5$ of them are downsample conv layers) and the decoder consists of $5$ ResBlocks~\cite{resnet}. Both are applied with spectral normalization~\cite{miyato2018spectral}. $I_{masked}$, $S_{masked}$, $M$ are concatenated as inputs and are encoded through the CNN encoder. Then the latent vector is forwarded to the decoder, which generates a extended segmentation map $S_{gen}$:
\begin{alignat}{2}
    S_{gen} = \mathbf{G_{seg}}(I_{masked}, S_{masked}, M) \\
    S_{out} = (1-M) \odot S_{gen} + S_{masked}.
\end{alignat}

The discriminator $\mathbf{D_{seg}}$ is used to discriminate $S_{out}$ from $S_{orig}$. Additionally, we also feed the mask $M$ and masked image $I_{masked}$ into $\mathbf{D_{seg}}$ to help it in the adversarial game. We adopt the multi-scale discriminator proposed in ~\cite{wang2018high} to improve the quality of image generated and avoid repetitive structures. 

To train the model well, a cross-entropy loss and an adversarial loss are introduced. The cross-entropy loss $\mathcal{L}_{ce}$ is applied between $S_{gen}$ and $S_{orig}$ to ensure the consistency between the known region and the extended region.
For adversarial loss we use hinge loss~\cite{hinge_loss} to stabilize training:
\begin{align}
    \mathcal{L}_{seg, D} &= \mathbb{E}[\max(0, 1-\mathbf{D_{seg}}(S_{orig}, I_{masked}))+ \nonumber \\ 
    & \max(0, 1+\mathbf{D_{seg}}(S_{out}, I_{masked}))] \\
    \mathcal{L}_{seg, G} &= \mathbb{E}[-\mathbf{D_{seg}}(S_{out},  I_{masked}))].
\end{align}
The two loss functions are combined with $\lambda_{ce}$:
\begin{align}
    \mathcal{L}_{seg} = \lambda_{ce} \mathcal{L}_{ce}(S_{gen}, S_{orig}) +\mathcal{L}_{adv}^{seg}.
    \label{eq:loss1}
\end{align}

After this stage, full segmentation map $S_{out}$ containing semantic and layout clues has been generated. 

\subsection{Semantic Image Synthesis}

In this stage, masked image $I_{masked}$, full segmentation map $S_{out}$ and binary mask $M$ are used to generate extended image $I_{out}$, which can be treated as a conditional semantic image synthesis task: given the segmentation map $S_{out}$, synthesize real image consistent with the masked image $I_{masked}$. The biggest difference from traditional semantic image synthesis tasks is that the synthesised image is required to be consistent with the masked image. 

We find that spatially-adaptive de-normalization (SPADE)~\cite{park2019semantic} shows a powerful ability when involving semantics to guide image outpainting. $I_{masked}$, $S_{out}$ and $M$ are concatenated as the input of the generator $\mathbf{G_{img}}$, which is similar to $\mathbf{G_{seg}}$, but equipped with SPADE modules in each block of its decoder. Specifically, the SPADE module exploits $S_{out}$ and $M$ as semantics to guide the image synthesis. Formally,
\begin{align}
    I_{gen} &= \mathbf{G_{img}}(I_{masked}, S_{out}, M)\\
    I_{out} &= (1-M) \odot I_{gen} + I_{masked}.
\end{align}

To train model $\mathbf{G_{img}}$ well, there are two main issues. On the one hand, the new generated part is supposed to be meaningful. On the other hand, the part near boundary ought to be natural, smooth and undistinguishable.
We use L$1$-distance $\mathcal{L}_{L1}$  and perceptual loss~\cite{NIPS2016_generating, Johnson2016Perceptual} $\mathcal{L}_{perc}$ between $I_{gen}$ and $I_{orig}$ to ensure the semantic consistency and style consistency at the boundary:
\begin{align}
    \mathcal{L}_{perc} &=  \sum_{i=1}^{N} \lambda_{perc}^i||\phi_i(I_{gen})-\phi_i(I_{orig})||_1
\end{align}
where $N=5$,  $\lambda_{prc}^i = 1/2^i$ and $\phi_i$ is the feature after the $i$-th ReLU layer of pre-trained VGG-19~\cite{2014Very} network. And we use adversarial hinge loss to make $\mathbf{G_{img}}$ generating realistic results:
\begin{align}
    \mathcal{L}_{img, D} &= \mathbb{E}[\max(0, 1-\mathbf{D_{img}}(I_{orig}, S_{orig}))+ \nonumber \\ 
    & \max(0, 1+\mathbf{D_{img}}(I_{out}, S_{orig}))] \\
    \mathcal{L}_{img, G} &= \mathbb{E}[-\mathbf{D_{img}}(S_{orig},  I_{masked}))].
\end{align}
We adopt the same multi-scale discriminator in semantic segmentation outpainting stage for $\mathbf{D_{img}}$. The final loss function $\mathcal{L}_{img}$ is a combination of the above:
\begin{align}
    \mathcal{L}_{img} = \lambda_{perc}\mathcal{L}_{perc} + \lambda_{L1}\mathcal{L}_{L1} + \mathcal{L}_{adv}^{img}.
    \label{eq:loss2}
\end{align}

\section{Experiments}

\subsection{Datasets} 
To evaluate our method, we perform experiments on two complex datasets:

\Paragraph{ADE20K}\cite{zhou2017scene} contains 20,210 training images and 2,000 validation images annotated with 150 semantic categories included. This dataset covers a vast variety of complex scenes and objects, both indoor and outdoor, which is much more challenging than scenery datasets used before. Due to the difficulty of extending multiple complex objects in intricate scenes, no image outpainting work has been done on this dataset, to the best of our knowledge.

\Paragraph{Cityscapes}\cite{Cordts2016Cityscapes} contains 2,975 training images and 500 validation images recorded in street scenes from 50 different cities in Germany. The images are finely annotated with 34 semantic categories. This dataset is easier than ADE20K, and some previous work~\cite{wang2019wide} has performed experiments on this dataset.

We adopt settings similar to~\cite{teterwak2019boundless}: aiming to fill in masked out image content where the rightmost 25\% and 50\% of pixels in the image are respectively masked. Specifically, for ADE20K~\cite{zhou2017scene} dataset, the input image to be masked is $256 \times 256$. For Cityscapes~\cite{Cordts2016Cityscapes} dataset, we crop the input 1:2 street-view image from middle-line to two 1:1 images, and flip the left one horizontally, which together compose the training set. During inference, we concatenate two extended images back to one 1:2 images (flipping the left image) to get the final extended image. This setting is reasonable because street-view images are symmetric and collected from the perspective of driving position.

During training, we use two common data augmentation methods for both datasets: random crop and random flip. Input images are resized to $286 \times 286$ first and randomly cropped to $256 \times 256$. We flip images horizontally with a probability of 0.5. Our models are trained on training set and all reported results are evaluated on validation set, following the official train/val split. 

\subsection{Implementation Details}
\label{sec:baseline}

We implement two alternatives of our method for ablation study. The first baseline is a one-stage and end-to-end model which shares similar structures as $\mathbf{G_{seg}}$ without the input of segmentation maps (\textbf{NoSeg}). The second one is similar to the first one while concatenating masked segmentation map with masked image as input (\textbf{SegConcat}). 

We use pre-trained ResNeSt-200~\cite{zhang2020resnest} and HRNet-OCR~\cite{tao2020hierarchical} for $\mathbf{SegNet}$. In training phase of image synthesis stage, we use $S_{orig}$ instead of $S_{out}$ since there is no extended RGB image to match $S_{out}$. We use $I_{orig}$ and $S_{orig}$ pairs to conduct supervised learning and train $\mathbf{G_{img}}$ to learn the mapping relationship between semantic segmentation maps and RGB pixels.We set $\lambda_{ce}=100$, $\lambda_{perc}=10$ and $\lambda_{L1}=100$ as trade-off parameters. We perform $300$ epochs of training for each model and dataset. Leaky-ReLU is used as the activation function both in generators and discriminators. Learning rates for generators and discriminators are initially set to $0.0001$ and $0.0004$ respectively, which decay linearly to $0$ after $200$ epochs. In addition, we apply spectral norm~\cite{miyato2018spectral}, synchronized BatchNorm~\cite{ioffe2015batch} and use Adam optimizer ~\cite{kingma2014adam} with $\beta_1=0,\beta_2=0.9$ to minimize the loss functions \ref{eq:loss1} and \ref{eq:loss2}. All models are implemented with Pytorch.

For comparison, we train state-of-the-art image outpainting models to compare with ours: Boundless~\cite{teterwak2019boundless} and NS-Outpainting~\cite{yang2019very}. We re-implement Boundless with PyTorch as there are no official training codes. For NS-Outpainting, we adopt their official implements and modify the data loader and input/output ratio of RCT module to adapt to our task settings. All experiments are conducted on NVIDIA 32GB V100 GPUs. 

\begin{table}[h!]
\begin{subtable}[h]{\columnwidth}
\centering
\begin{tabular}{@{}lcc@{}}
\toprule
Method              & FID (25\%)                     & FID (50\%)                       \\ \midrule
Boundless~\cite{teterwak2019boundless}           & 12.82                        & 38.51                           \\
NS-Outpainting  ~\cite{yang2019very} & 14.42                          & 42.29                           \\
NoSeg               & 11.69                          & 34.95                           \\
SegConcat           & 10.11                              & 35.18                               \\
Ours                & \textbf{8.26}                  & \textbf{27.67}                 \\ \bottomrule
\end{tabular}
\caption{Results on ADE20K dataset.}
\label{table:ade20k}
\end{subtable}
\newline
\vspace{0.4cm}
\newline
\begin{subtable}[h]{\columnwidth}
\centering
\begin{tabular}{@{}lcc@{}}
\toprule
Method              & FID (25\%)                     & FID (50\%)                       \\ \midrule
Boundless~\cite{teterwak2019boundless}           & 22.44                          & 41.08                           \\
NS-Outpainting~\cite{yang2019very} & 20.04                          & 43.36                           \\
NoSeg               & 20.84                          & 45.11                           \\
SegConcat           & 19.81                              & 47.71                               \\
Ours                & \textbf{15.28}                  & \textbf{33.80}                 \\ \bottomrule
\end{tabular}
\caption{Results on Cityscapes dataset.}
\label{table:cityscapes}
\end{subtable}
\caption{FID results on ADE20K and Cityscapes validation sets. FID is computed on the whole validation set. We compare Boundless~\cite{teterwak2019boundless}, NS-Outpainting~\cite{yang2019very} and our methods. Some possible alternatives are compared for ablation study.}
\label{table:quantitative}
\end{table}

\begin{figure*}[p]
\centering
\centerline{\includegraphics[width=0.99\textwidth]{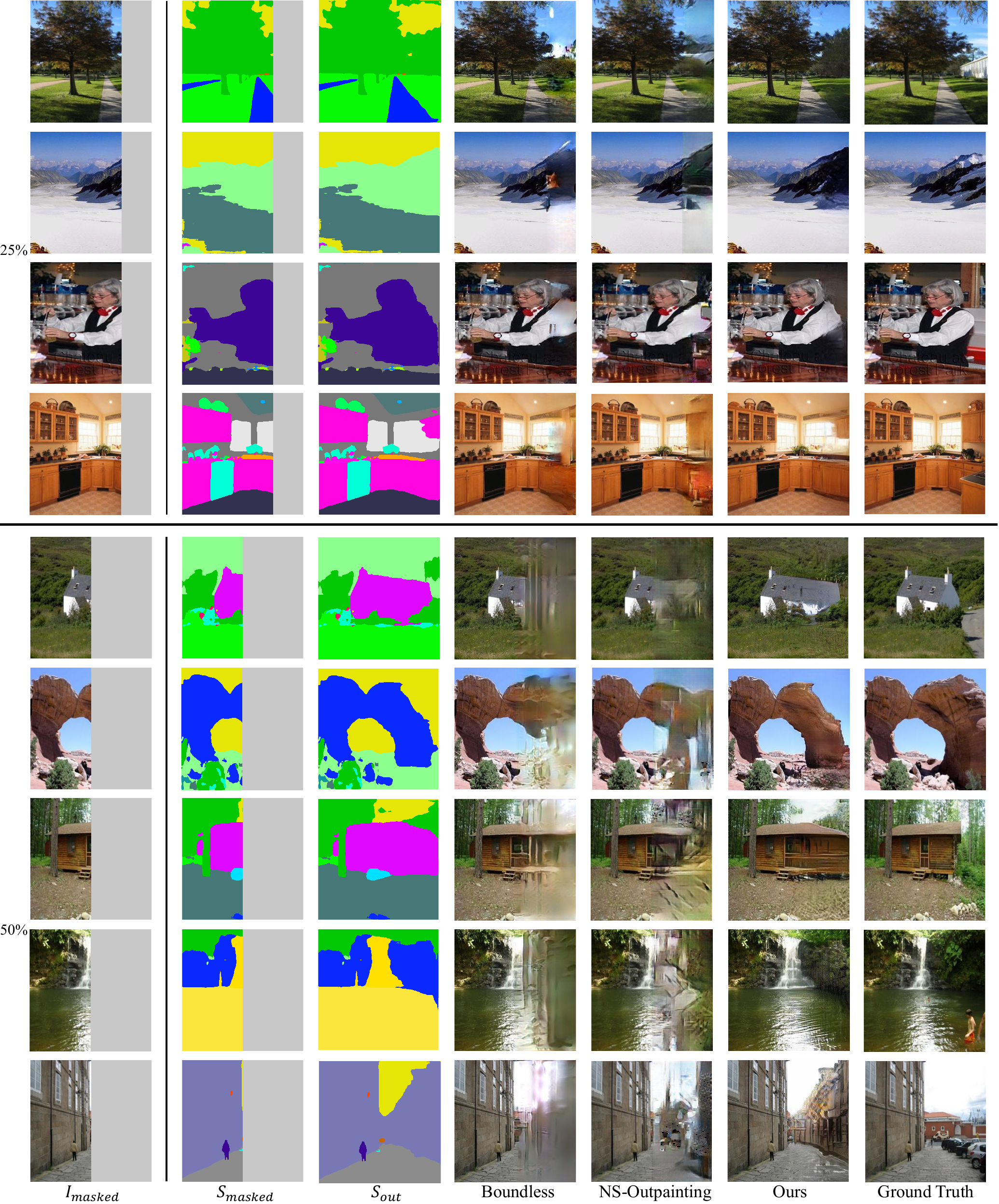}}
\caption{Qualitative results for proposed models on ADE20K dataset~\cite{zhou2017scene} of 25\% and 50\% masks. From left to right: input image $I_{masked}$, input image's segmentation map $S_{masked}$, extended segmentation map $S_{out}$,  Boundless~\cite{teterwak2019boundless}, NS-Outpainting~\cite{yang2019very}, our model and ground truth. Please zoom in the images for better view.}
\label{fig:comparison}
\end{figure*}

\begin{figure*}[t]
\centering
\centerline{\includegraphics[width=\textwidth]{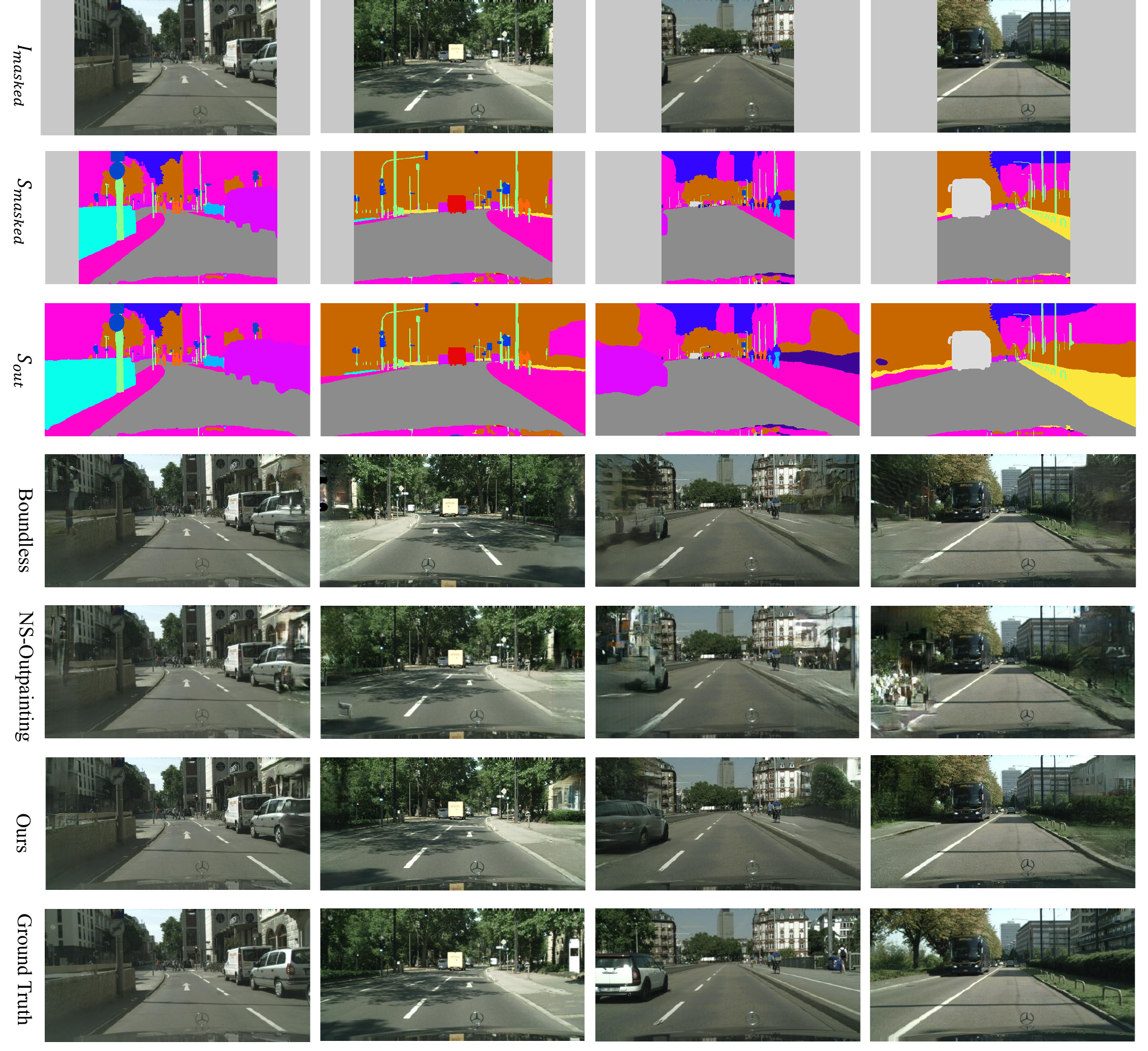}}
\caption{Qualitative results for proposed models on Cityscapes~\cite{Cordts2016Cityscapes} of 25\% and 50\% masks. From top to bottom: input image $I_{masked}$, input image's segmentation map $S_{masked}$, extended segmentation map $S_{out}$,  Boundless~\cite{teterwak2019boundless}, NS-Outpainting~\cite{yang2019very}, our model and ground truth. Please zoom in the images for better view.}
\label{fig:comparisoncity}
\end{figure*}

\subsection{Comparison with Previous Works}

\begin{table}[h]
\centering
\begin{tabular}{@{}cccc@{}}
\toprule
Datasets          & Boundless & NS-Outpaint & Ours \\ \midrule
ADE20K (25\%)     & 4.17\%         & 7.92\%              & \textbf{87.92\%}    \\
ADE20K (50\%)     & 5.42\%         & 2.50\%              & \textbf{92.08\%}    \\
Cityscapes (25\%) & 14.17\%         & 19.17\%              & \textbf{66.67\%}   \\
Cityscapes (50\%) & 15.42\%         & 6.25\%              & \textbf{78.33\%}    \\ \bottomrule
\end{tabular}
\caption{User study statistics. Each represents the percentage of cases where that method is the best, under the ratio setting.}
\label{table:userstudy}
\end{table}

\Paragraph{Quantitative Results} In order to quantitatively measure the quality of generated images, one universally acknowledged metric is Fréchet Inception Distance (FID)~\cite{NIPS2017_8a1d6947}, which measures the distance in feature space between generated and real images.

We compute FID scores on the whole validation sets of ADE20K~\cite{zhou2017scene} and Cityscapes~\cite{Cordts2016Cityscapes}, using two mask ratio (25\% and 50\%) settings. The results of two datasets are shown in Table~\ref{table:quantitative}. We can observe that our method significantly (more than 20\%) outperforms other methods on both datasets and both ratio settings, which proves the effectiveness of our framework.

Considering the limitation of FID, we also conduct a user study to evaluate human perception preference. We randomly select 40 images (20 from ADE20K and 20 from Cityscapes) and invite 24 participants who are not involved in this project to the user study. Each participant is asked to pick the best image of the three (Boundless, NS-Outpainting and ours). Images are shuffled in a blind random order to avoid guessing sequences of the three. Specific statistics are shown in Table~\ref{table:userstudy}, which indicates consistent preference of our method over other baseline methods. 

\Paragraph{Qualitative Results} Figure~\ref{fig:comparison} and Figure~\ref{fig:comparisoncity} give qualitative comparison of various outpainting methods on ADE20K and Cityscapes dataset respectively. We can see that in complex scenarios, especially ADE20K, our method generates clearer and more reasonable extended images compared to others. Other methods usually degenerate when objects stand on the boundary to extend, while ours works well for exploiting semantic information. Specifically, Boundless~\cite{teterwak2019boundless} and NS-Outpainting~\cite{yang2019very} both tend to extend boundary pixels horizontally. This seems helpful for natural scenery images but usually leads to fake artifacts for images containing complex objects. 

\Paragraph{Summary} Both qualitative and quantitative results show that our method outperforms others by a large margin, especially in complex scenes. Our extended semantic segmentation maps are geometrically reasonable and symmetric, consistent with the distribution of real images. In addition, there is a strong correlation between the extended image and the extended segmentation map, which coincides with our motivation of decomposing the outpainting task, proving the effectiveness of our two-stage method.

\begin{figure}
\centering
\includegraphics[width=\columnwidth]{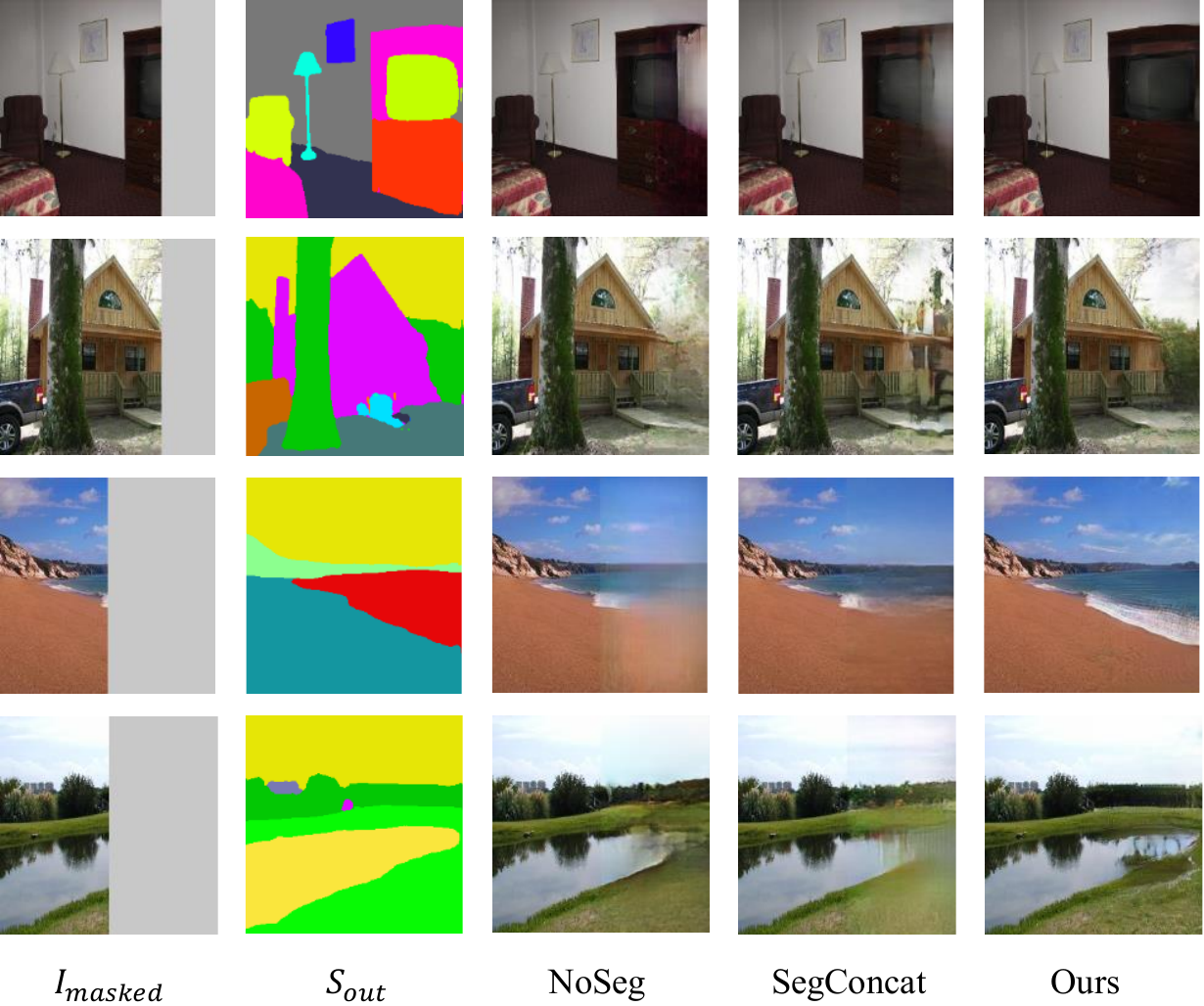}
\caption{Ablation results. Our two-stage method produces higher quality images with sharper boundaries and more consistent semantics.} \label{fig:ablation}
\end{figure}

\subsection{Ablation Studies and Discussion}

We conduct ablation studies on the effectiveness of semantic information and the necessariness of our two-stage strategy.

As mentioned in Section~\ref{sec:baseline}, we implement two weaker models (NoSeg, SegConcat) to prove the necessity of our proposed method for outpainting. We also report their quantitative results in Table~\ref{table:quantitative}. Not surprisingly, ours gets the best FID value.
SegConcat performs better than NoSeg at the mask of 25\% ratio, but is even slightly worse than NoSeg at a larger ratio 50\%. This is due to the limited guidance provided by partial semantic segmentation maps, which only helps on small-ratio outpainting. 

We also give a qualitative comparison of the three methods in Figure~\ref{fig:ablation}. It shows that the NoSeg method can hardly control the edge of objects well, generating fuzzy and low-quality images. Especially when there are objects at the boundary, it is impossible to achieve semantic consistency without any guidance. The other two methods both leverage semantic segmentation information for image outpainting. For SegConcat model, it involves masked semantic segmentation map as part of input, relying on model itself to extend semantic information implicitly. But as for ours, we explicitly exploit the extended semantic segmentation map and generate images under the guidance of it. Obviously, our method produces compelling extended semantic layouts and semantically consistent extended images.

Our experiments prove that involving semantic segmentation map facilitates the outpainting task a lot and the decomposition of the task is effective.

\begin{figure}
\centering
\includegraphics[width=\columnwidth]{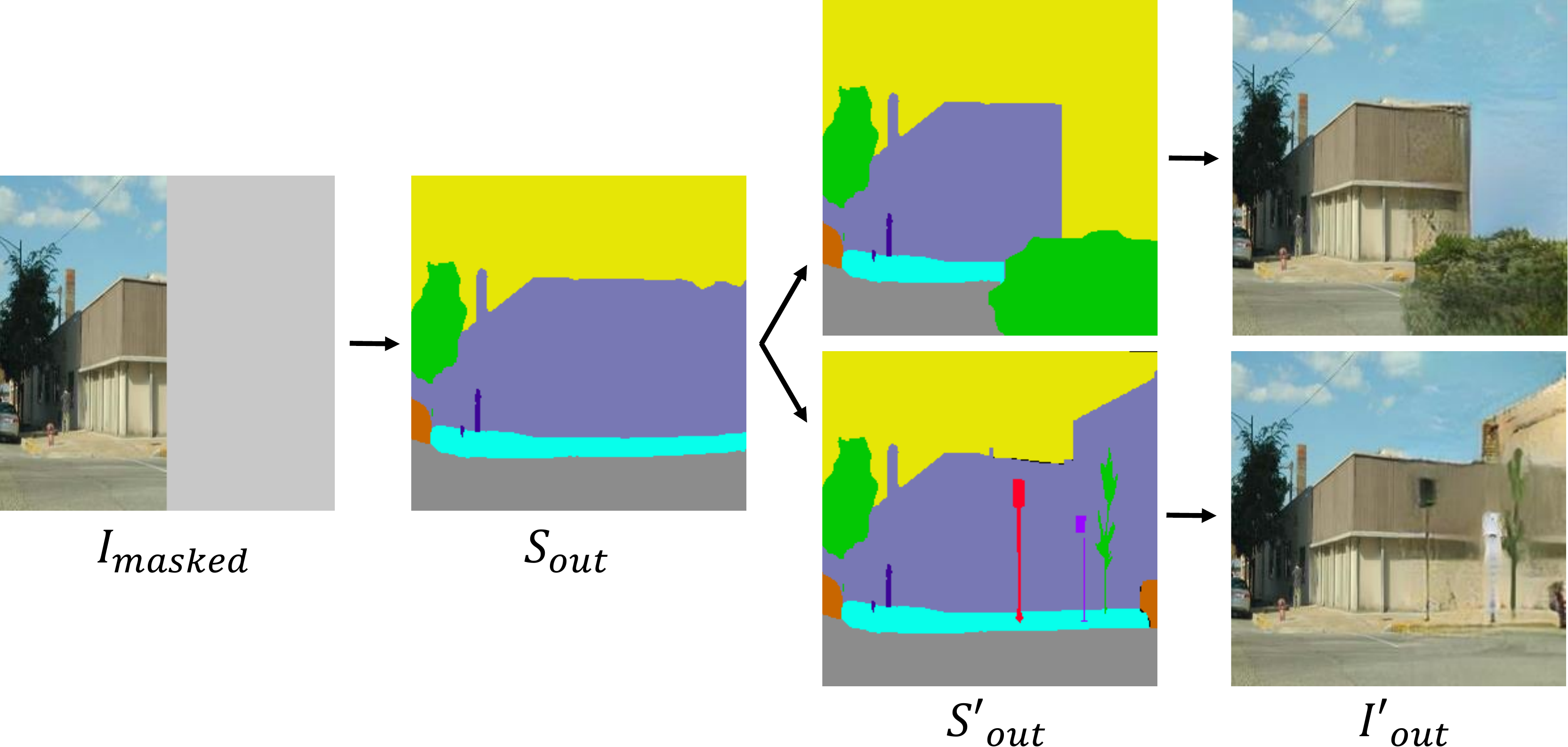}
\caption{Editable outpainting: designers could conveniently edit extended semantic layout as they wish based on $S_{out}$, generating diverse convincing and reasonable outpainting images $I^{\prime}_{out}$.} \label{fig:multi}
\end{figure}

\subsection{Applications and Limitations}

Image outpainting could be widely used in applications such as automatic image editing (Figure~\ref{fig:intro}). In addition, since we decompose the image outpainting task to two stages and exploit semantic layout explicitly, this naturally gives us the chance to manually adjust outpainting results, which means we could conveniently modify extended semantic layout to generate diverse outpainting images, as shown in Figure~\ref{fig:multi}. This capacity is beneficial for interactive image editing tools but could not be easily adopted by previous methods.

Although our method performs obviously better than related work in complex scene datasets, it still suffers from some structural objects such as humans. This may be caused by the uneven distribution of the objects in complex scenes and our dataset is not so big ($20$k) compared to others ($1$M), leading to poor knowledge of structure and texture information. 


\section{Conclusion}

In this paper, we propose a two-stage image outpainting framework to generate high-quality extended images. The main idea is to disassemble the outpainting task into two parts: semantic segmentation outpainting and semantic image synthesis. The former focuses on semantic and structural consistency at the boundary while the latter ensures realistic texture, color and lighting. 
Our method behaves significantly better when handling complex scenarios, outperforming state-of-the-art methods by a large margin.

{\small
\bibliographystyle{ieee_fullname}

}

\clearpage

\appendix

\section{Additional Implementation Details}
We give a detailed description of our network structure. Table~\ref{table:discrim} shows the multi-scale discriminator we use which is proposed in ~\cite{wang2018high}.  $\mathbf{G_{img}}$ and $\mathbf{G_{seg}}$ both follow an encoder-decoder structure. $\mathbf{G_{img}}$ employs the same encoder with $\mathbf{G_{seg}}$, while equipped with SPADE modules in each block of its decoder. The common encoder structure and the decoder structures are shown in Table~\ref{table:encoder}, Table~\ref{table:decoder0} and Table~\ref{table:decoder1} respectively.

\begin{table}[!h]
\centering
\begin{tabular}{c|c|c|c}  
\hline  
\textbf{Type} & \textbf{Kernel Size} & \textbf{Stride} & \textbf{\# Channels} \\\hline  
 Conv.   & $4\times4$  & $2$ & $64$          \\\hline  
 Conv.   & $4\times4$  & $2$ & $128$          \\\hline  
 Conv.   & $4\times4$  & $2$ & $256$          \\\hline  
 Conv.   & $4\times4$  & $1$ & $512$          \\\hline  
 Conv.   & $4\times4$  & $1$ & $1$          \\\hline  
\end{tabular}  
\caption{Architecture details of discriminators. We apply synchronized BatchNorm~\cite{ioffe2015batch} and Spectral Norm~\cite{miyato2018spectral} to the convolutional layers except the first and last layers. We use LeakyReLU as activation function for all convolutional layers.}
\label{table:discrim}
\end{table}

\begin{table}[!h]
\centering
\begin{tabular}{c|c|c|c}  
\hline  
\textbf{Type} & \textbf{Kernel Size} & \textbf{Stride} & \textbf{\# Channels} \\\hline  
 Conv.   & $7\times7$  & $1$ & $64$          \\\hline  
Conv.    &  $3\times3$ & 1  &  $128$           \\\hline  
Conv.    & $3\times3$ & 2  &  $128$           \\\hline  
Conv.    &  $3\times3$ & 1  &  $256$           \\\hline  
Conv.    & $3\times3$ & 2  &  $256$           \\\hline  
Conv.    &  $3\times3$ & 1  &  $512$           \\\hline  
Conv.    &  $3\times3$ & 2  &  $512$           \\\hline  
Conv.    & $3\times3$ & 1  &  $1024$           \\\hline  
Conv.    &  $3\times3$ & 2  &  $1024$           \\\hline  
Conv.    & $3\times3$ & 1  &  $1024$           \\\hline  
Conv.    & $3\times3$ & 2  &  $1024$           \\\hline  
\end{tabular}  
\caption{Architecture details of encoder. We apply SyncBatchNorm~\cite{ioffe2015batch}, LeakyReLU and Spectral Norm~\cite{miyato2018spectral} to all the convolutional layers.}
\label{table:encoder}
\end{table}

\begin{table}[!h]
\centering
\begin{tabular}{c|c|c|c}  
\hline  
\textbf{Type} & \textbf{Kernel Size} & \textbf{Stride} & \textbf{\# Channels} \\\hline  
 ResnetBlock   & $3\times3$  & $1$ & $1024$          \\\hline  
 Upsample    &  None & None  &  $1024$           \\\hline  
ResnetBlock   & $3\times3$ & 1  &  $1024$           \\\hline  
ResnetBlock   & $3\times3$ & 1  &  $1024$           \\\hline  
 Upsample    &  None & None  &  $1024$           \\\hline  
ResnetBlock   &  $3\times3$ & 1  &  $512$           \\\hline  
 Upsample    &  None & None  &  $512$           \\\hline  
ResnetBlock   &  $3\times3$ & 1  &  $256$           \\\hline  
 Upsample    &  None & None  &  $256$           \\\hline  
ResnetBlock   &  $3\times3$ & 1  &  $128$           \\\hline  
 Upsample    &  None & None  &  $128$           \\\hline  
ResnetBlock   &  $3\times3$ & 1  &  $64$           \\\hline  
Conv.    & $3\times3$ & 1  &  $ \text{out\_nc}$           \\\hline  
\end{tabular}  
\caption{Architecture details of decoder in $\mathbf{G_{seg}}$. For upsample we use the ``nearest'' mode and set scale\_factor to 2. We don't apply any activation function after last convolutional layer and the decoder generates a semantic segmentation map}
\label{table:decoder0}
\end{table}

\begin{table}[!h]
\centering
\begin{tabular}{c|c|c|c}  
\hline  
\textbf{Type} & \textbf{Kernel Size} & \textbf{Stride} & \textbf{\# Channels} \\\hline  
 SPADEBlock   & $3\times3$  & $1$ & $1024$          \\\hline  
 Upsample    &  None & None  &  $1024$           \\\hline  
SPADEBlock   & $3\times3$ & 1  &  $1024$           \\\hline  
SPADEBlock   & $3\times3$ & 1  &  $1024$           \\\hline  
 Upsample    &  None & None  &  $1024$           \\\hline  
SPADEBlock   &  $3\times3$ & 1  &  $512$           \\\hline  
 Upsample    &  None & None  &  $512$           \\\hline  
SPADEBlock   &  $3\times3$ & 1  &  $256$           \\\hline  
 Upsample    &  None & None  &  $256$           \\\hline  
SPADEBlock   &  $3\times3$ & 1  &  $128$           \\\hline  
 Upsample    &  None & None  &  $128$           \\\hline  
SPADEBlock   &  $3\times3$ & 1  &  $64$           \\\hline  
Conv.    & $3\times3$ & 1  &  $\text{out\_nc}$           \\\hline  
\end{tabular}  
\caption{Architecture details of decoder in $\mathbf{G_{img}}$. For upsample we use the ``nearest'' mode and set scale\_factor to 2. We apply tanh after last convolutional layer and the decoder generates a real image. }
\label{table:decoder1}
\end{table}

\begin{figure*}[p]
\centering
\centerline{\includegraphics[width=0.97\textwidth]{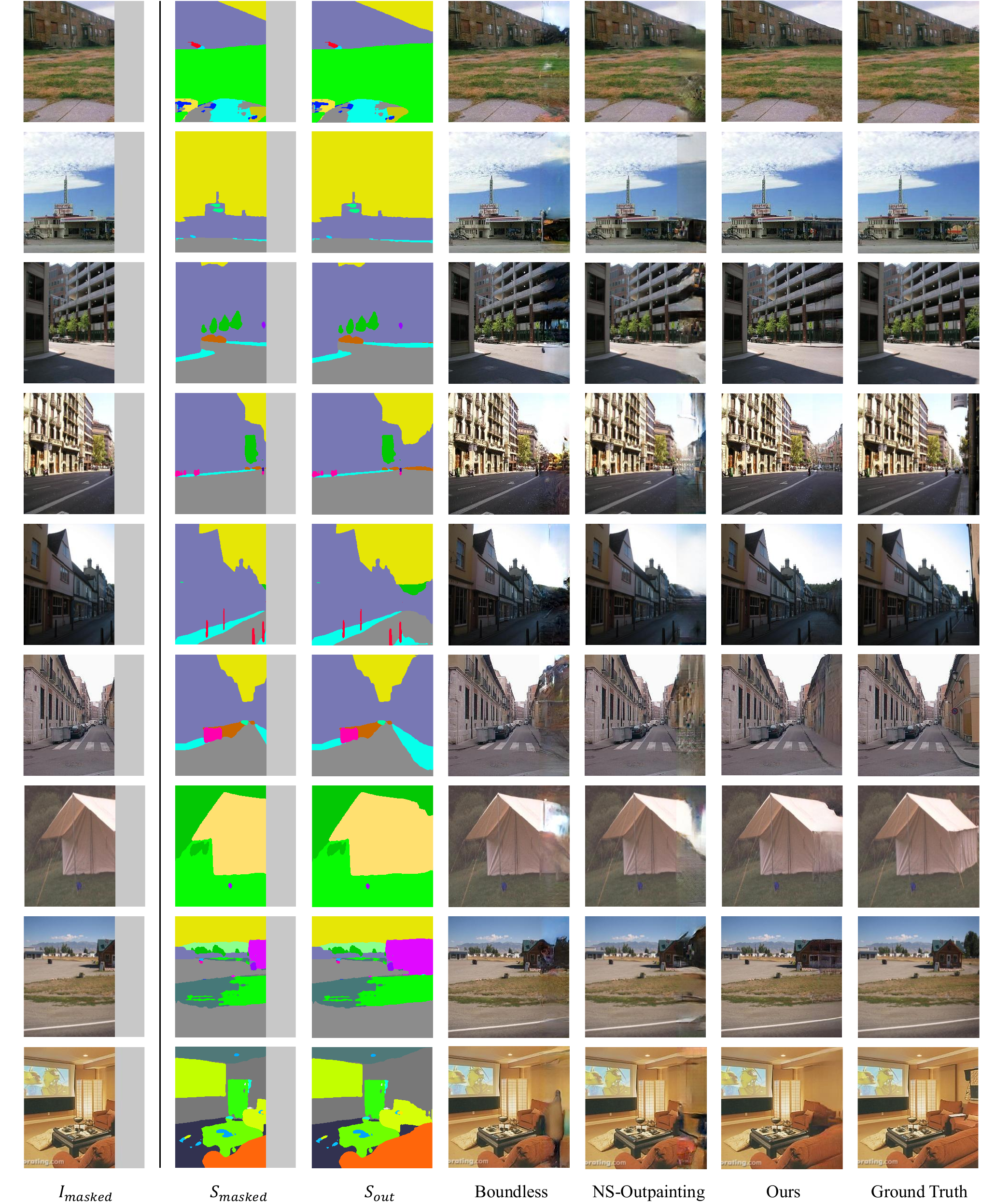}}
\caption{Additional qualitative results for proposed models on ADE20K dataset~\cite{zhou2017scene} of 25\% masks. From left to right: input image $I_{masked}$, input image's segmentation map $S_{masked}$, extended segmentation map $S_{out}$,  Boundless~\cite{teterwak2019boundless}, NS-Outpainting~\cite{yang2019very}, our model and ground truth. Please zoom in the images for better view.}
\label{fig:comparison25}
\end{figure*}

\begin{figure*}[p]
\centering
\centerline{\includegraphics[width=0.97\textwidth]{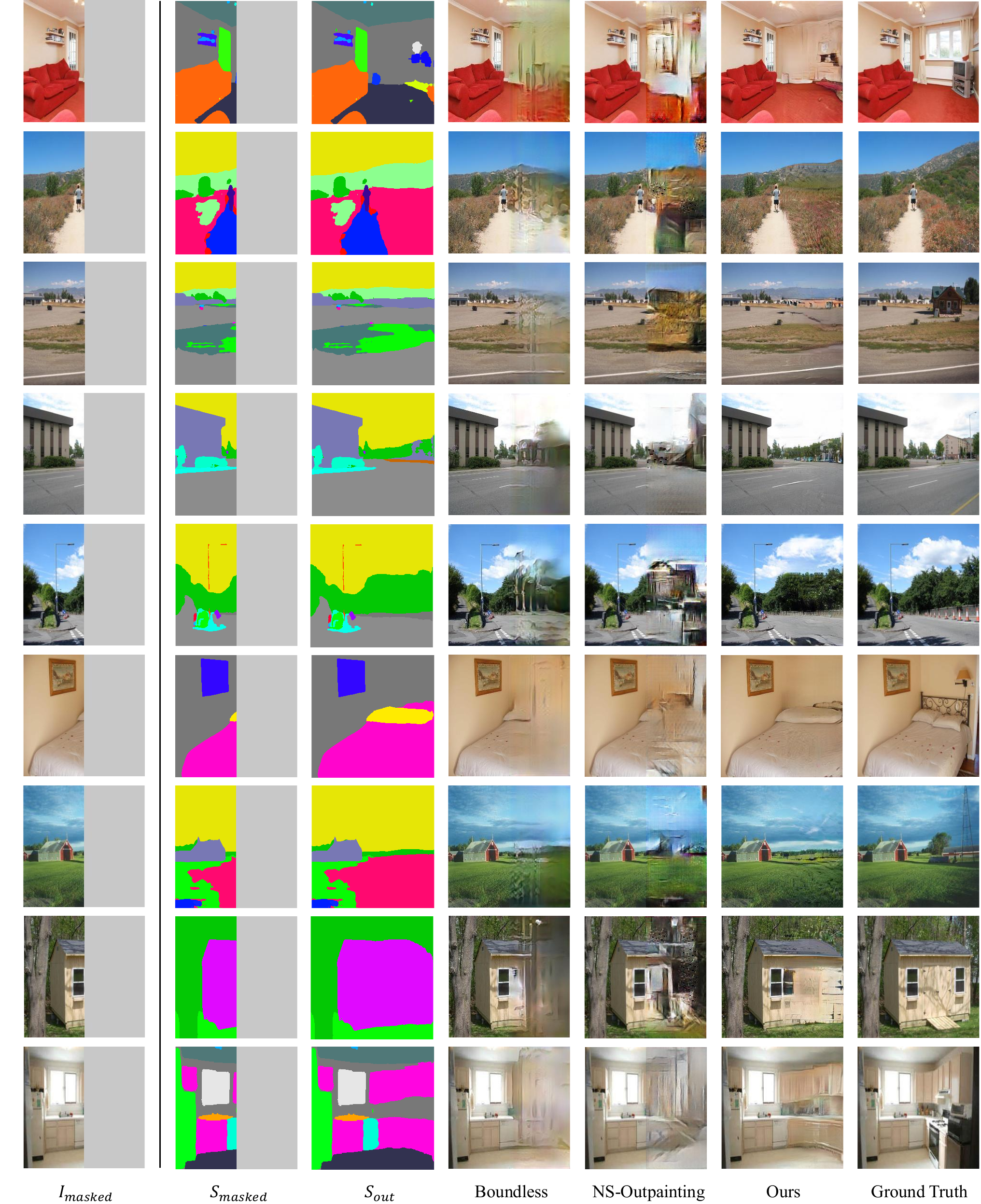}}
\caption{Additional qualitative results for proposed models on ADE20K dataset~\cite{zhou2017scene} of 50\% masks. From left to right: input image $I_{masked}$, input image's segmentation map $S_{masked}$, extended segmentation map $S_{out}$,  Boundless~\cite{teterwak2019boundless}, NS-Outpainting~\cite{yang2019very}, our model and ground truth. Please zoom in the images for better view.}
\label{fig:comparison50}
\end{figure*}

\begin{figure*}[!h]
\centering
\centerline{\includegraphics[width=\textwidth]{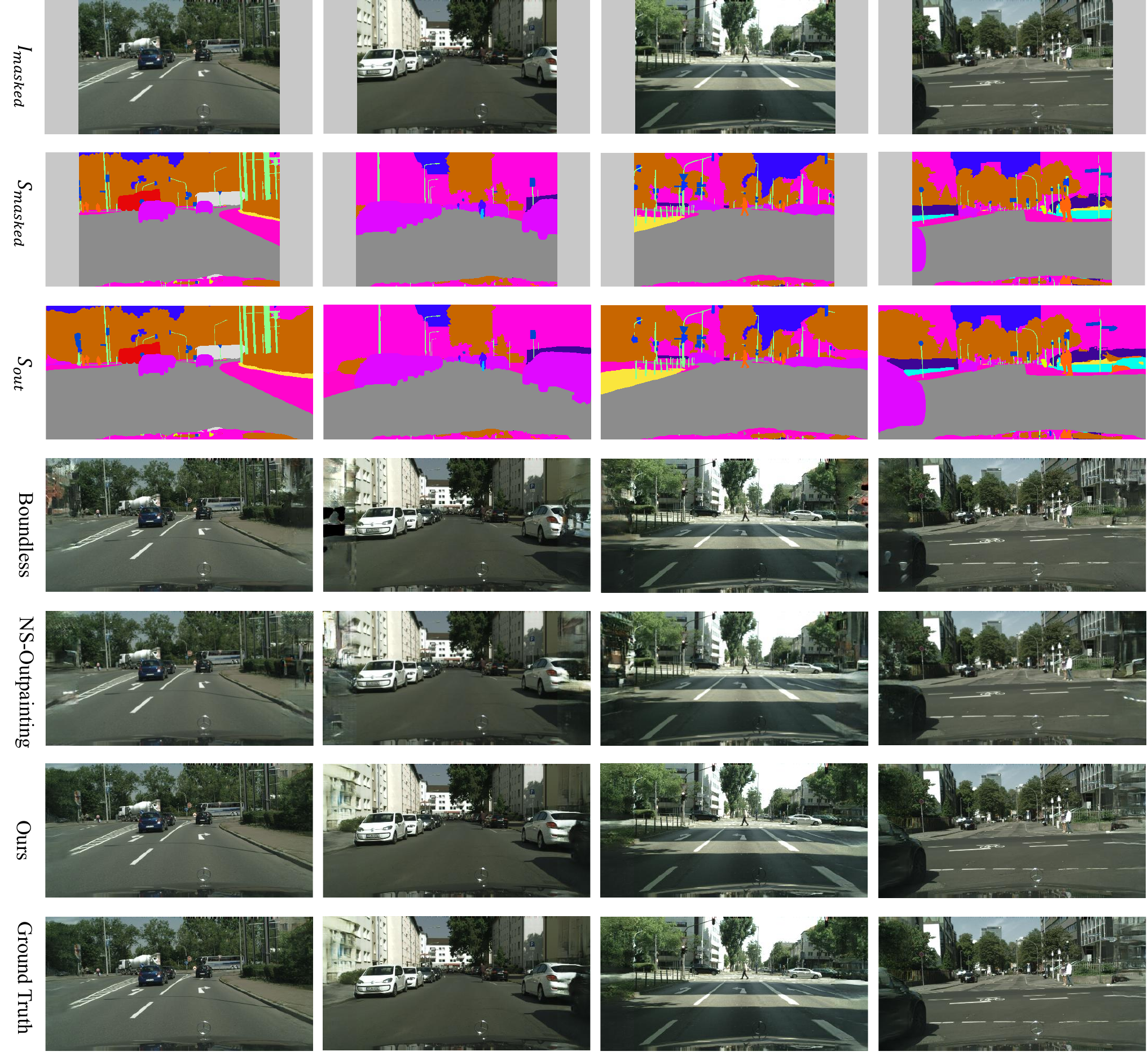}}
\caption{Additional qualitative results for proposed models on Cityscapes~\cite{Cordts2016Cityscapes} of 25\% masks. From top to bottom: input image $I_{masked}$, input image's segmentation map $S_{masked}$, extended segmentation map $S_{out}$,  Boundless~\cite{teterwak2019boundless}, NS-Outpainting~\cite{yang2019very}, our model and ground truth. Please zoom in the images for better view.}
\label{fig:city25}
\end{figure*}

\begin{figure*}[!h]
\centering
\centerline{\includegraphics[width=\textwidth]{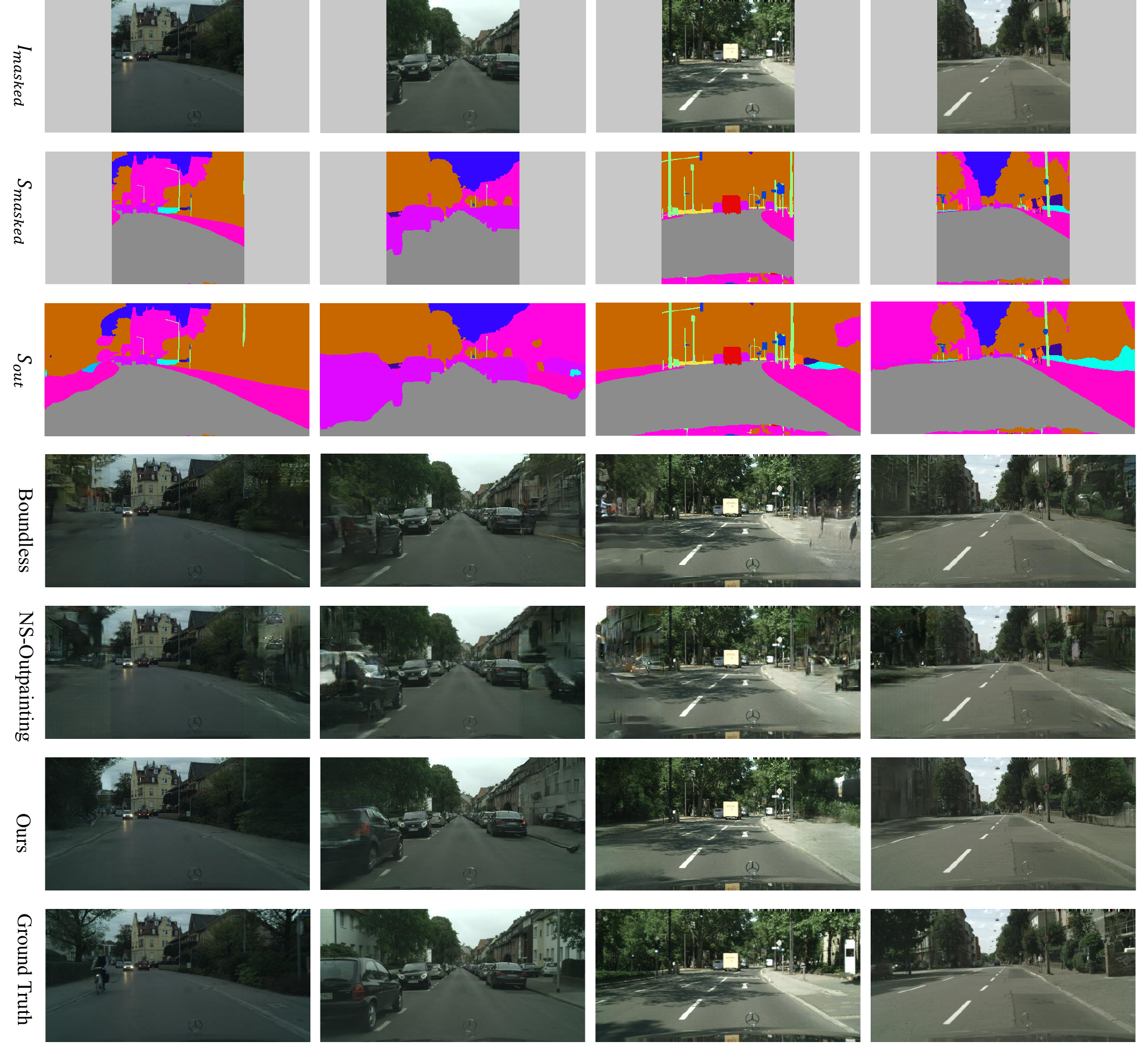}}
\caption{Additional qualitative results for proposed models on Cityscapes~\cite{Cordts2016Cityscapes} of 50\% masks. From top to bottom: input image $I_{masked}$, input image's segmentation map $S_{masked}$, extended segmentation map $S_{out}$,  Boundless~\cite{teterwak2019boundless}, NS-Outpainting~\cite{yang2019very}, our model and ground truth. Please zoom in the images for better view.}
\label{fig:city50}
\end{figure*}

\begin{figure*}[!h]
\centering
\centerline{\includegraphics[width=0.97\textwidth]{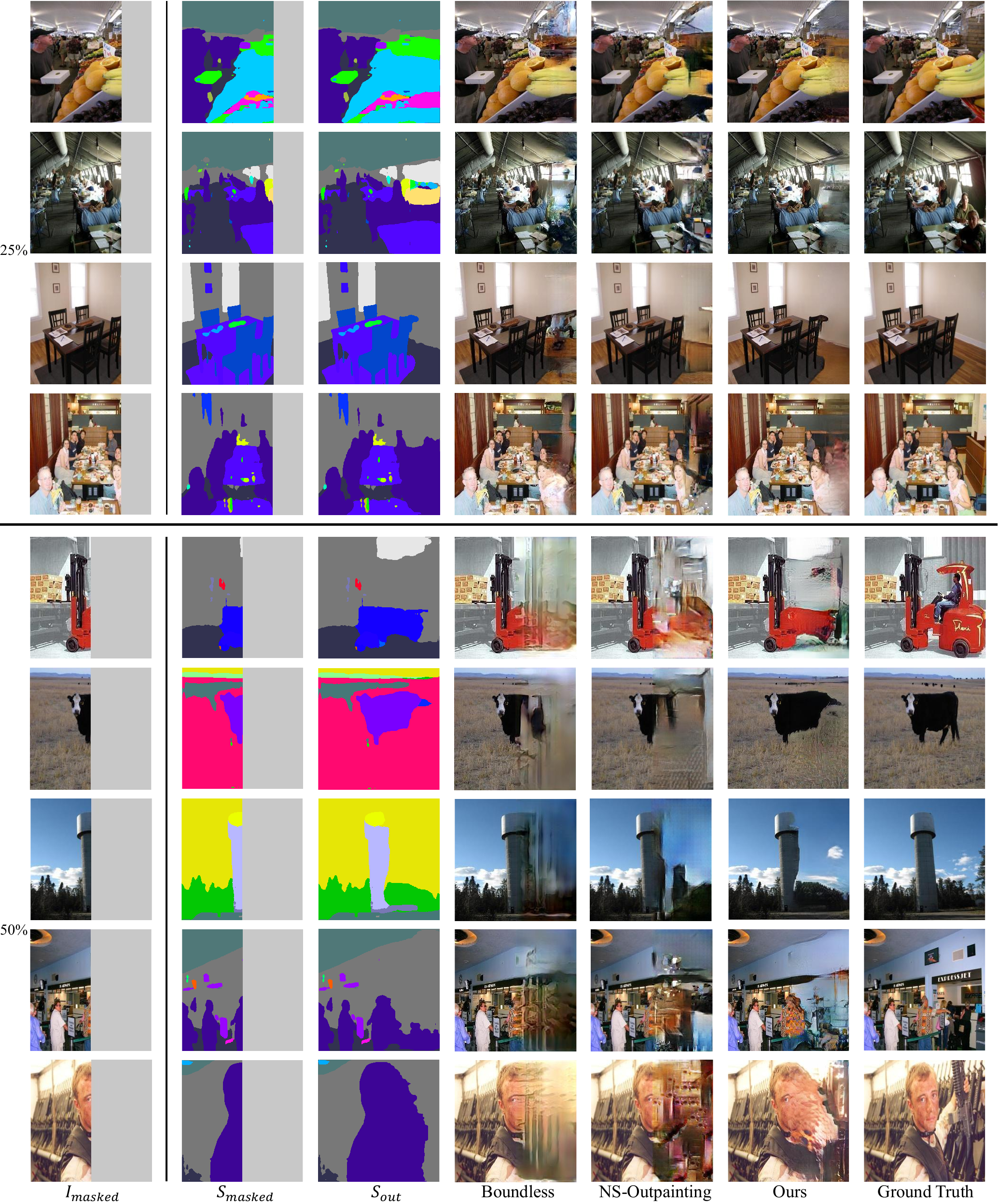}}
\caption{Some bad cases for proposed models on ADE20K dataset~\cite{zhou2017scene} of 25\% and 50\% masks. From left to right: input image $I_{masked}$, input image's segmentation map $S_{masked}$, extended segmentation map $S_{out}$,  Boundless~\cite{teterwak2019boundless}, NS-Outpainting~\cite{yang2019very}, our model and ground truth. Please zoom in the images for better view.}
\label{fig:bad}
\end{figure*}

\section{Additional Qualitative Results}
Figure~\ref{fig:comparison25}, Figure~\ref{fig:comparison50} and  Figure~\ref{fig:city25}, Figure~\ref{fig:city50} give additional qualitative comparisons of various outpainting methods under two mask ratio (25\% and 50\%) settings on ADE20K~\cite{zhou2017scene} and Cityscapes~\cite{Cordts2016Cityscapes} dataset respectively. 

\section{Bad Cases}

Our model still suffers from some structural objects such as vehicles and humans. 
We show some bad cases in Figure~\ref{fig:bad} and this is what we are to address in our future work.

\end{document}